\documentclass[runningheads]{llncs}
\usepackage{graphicx}
\usepackage{akkmathset}

\usepackage{algorithm}
\usepackage{algorithmic}
\usepackage{latexsym}
\usepackage{cite}
\usepackage{enumitem}
\usepackage{amssymb,mathrsfs}
\usepackage{booktabs}
\usepackage{microtype}
\usepackage{siunitx}
\usepackage{dcolumn}
\usepackage{colortbl}
\usepackage{subfigure}
\usepackage{float}
\usepackage{color}
\usepackage[dvipsnames]{xcolor}
\usepackage{amsmath}
\usepackage{booktabs}
\usepackage{multirow}

\begin{document}
\title{AdaFamily: A family of Adam-like \\adaptive gradient methods }
%
%
\author{Hannes Fassold}
%
%

\institute{DIGITAL - Institute for Information and Communication Technologies\\JOANNEUM RESEARCH, Graz, Austria \\ \email{hannes.fassold@joanneum.at}}

\maketitle              
\begin{abstract}
We propose \emph{AdaFamily}, a novel  method for training deep neural networks. It is a family of adaptive gradient methods and can be interpreted as sort of a \emph{blend} of the optimization algorithms Adam, AdaBelief and AdaMomentum. We perform experiments on standard datasets for image classification, demonstrating that our proposed method outperforms these algorithms.
\end{abstract}
\section{Introduction}

Adaptive gradient methods, especially the \emph{Adam} algorithm \cite{Kingma2015AdamAM} are nowadays the standard for training deep neural networks. This is because they are much easier to handle (less sensitive to weight initialization and hyperparameters) compared with mini-batch stochastic gradient descent  \cite{Dekel2012} (SGD). This can be attributed to two key properties of the Adam algorithm: \emph{adaptive learning rate} and \emph{momentum term}. Via the \emph{adaptive learning rate}, the algorithm computes an individual learning rate for each parameter of the neural network model - in contrast to SGD which keeps a fixed learning rate for all parameters of the model. This is especially helpful for neural networks with many different types of layers, with the parameters in each layer typically having a different "typical" value range. On the other side, via the \emph{momentum term} it can keep progress even when encountering regions of the loss function landscape with high curvature, which resemble long and narrow ravines (mathematically speaking, these are regions where the Hessian matrix has a high condition number).

Since the introduction of the Adam algorithm, many variations of the algorithm have been proposed (see Table 2 in the survey paper \cite{Schmidt2021}). Although many of them might have marginal benefits in practice \cite{Schmidt2021}, a few of them seem to be able to really provide an improvement of the training process when compared to Adam. Specifically, the work \cite{Loshchilov2019} demonstrates that $L_{2}$ regularization and weight decay are \emph{not} equivalent for adaptive gradient methods. Based on their analysis, they propose the \emph{AdamW} algorithm, which decouples the weight decay from the gradient-based update. The \emph{AdaBelief} algorithm \cite{Zhuang2020} modifies the calculation of the scaling term $v_t$ (corresponding to the denominator in the weight update step of the Adam algorithm) in a way which takes into account the "belief" in the gradient. Experiments with a variety of models (CNN, LSTM, GAN) show that this modification improves the performance of the training. Furthermore, recently the \emph{AdaMomentum} algorithm \cite{Wang2021} was proposed. It also modifies the calculation of the scaling term $v_t$ by replacing the gradient with its exponential moving average (EMA). They also propose to move the addition of the constant $\epsilon$ (to prevent for the division by zero) to a different term and justify why this is advantageous.

AdaBelief and AdaMomentum differ from Adam in only a few places, most prominently in the term corresponding for the scaling term $v_t$. The scaling term $v_t$ can be interpreted also as a diagonal preconditioner \cite{Qu2020} which is applied (left-multiplied) to the gradient prior to the weight update, so these algorithms differ mainly by the way how the preconditioner is calculated. Inspired by this observation and by the question whether we can "blend" these algorithms together in a useful way, we propose \emph{AdaFamily}, a family of Adam-like algorithms parametrized by a hyperparameter $\mu$ lying in range $[0, 1]$. All variants of the AdaFamily algorithm can be seen in a certain way as a \emph{blend} of the Adam/AdamW, AdaBelief and AdaMomentum algorithm.

In section \ref{sec:algorithm}, a description of the AdaFamily algorithm is given. In section \ref{sec:experiments}, we describe the experiments and show that by blending the original algorithms together, the AdaFamily algorithm is able to outperform them on standard image classification training tasks. Finally, section \ref{sec:conclusion} concludes our work.

\section{AdaFamily algorithm}
\label{sec:algorithm}

In the following, we formally define the AdaFamily optimization algorithm for training a deep neural network. We first describe the used notation and provide the pseudo code of the AdaFamily algorithm. After that, we elaborate on how AdaFamily is related to Adam/AdamW, AdaBelief and AdaMomentum and how the hyperparameter $\mu$ can be interpreted.

Regarding \emph{notation}, we denote with $t$ the current iteration (current step) in the training process. $\theta \in \RR^d$ denotes the model parameter and $f(\theta) \in \RR$ denotes the loss function. We further use $\theta_t$ to denote the parameter at step $t$ and $f_t$ to denote the noisy realization of $f$ at time $t$ due to the stochastic mini-batch mechanism. The gradient of $f_t$ is denoted as $g_t$, and $\alpha$ is the stepsize (learning rate). $m_t$ represents the exponential moving average of the gradient, whereas $v_t$ corresponds to the scaling term (preconditioning term). $\epsilon$ is a small constant number added in adaptive gradient methods to refrain the denominator from being too close to zero. $\beta_1, \beta_2$ are the decaying parameter in the EMA formulation of $m_t$ and $v_t$ correspondingly.  For any vectors $a,b \in \RR^d$, we employ $\sqrt{a}, a^2, |a|, a/b$ for \emph{elementwise} square root, square, absolute value or division respectively. 

The overall workflow of the AdaFamily algorithm is given now as pseudo code in Algorithm \ref{alg:adafamily}. One can see that the main difference to Adam and its variations is the term (we will denote it by $S$) which is element-wise squared in the computation of the preconditioner $v_t$. The variable $c$ can be seen as a normalization factor, where $c(\mu)$ is a slightly modified triangle function which returns $1.0$ for $\mu=0.0$ or $\mu=1.0$ and  $2.0$ for $\mu=0.5$. 

\begin{algorithm}[t]
\caption{AdaFamily algorithm. \\The main differences to the Adam algorithm are marked in {\color{NavyBlue}navy blue}.}
\label{alg:adafamily}

\begin{algorithmic}

{\color{NavyBlue}\REQUIRE $\mu$: hyperparameter in range $[0, 1]$}
\REQUIRE $f(\theta)$: objective function (loss function) with model parameters $\theta$
\REQUIRE $\alpha$: stepsize (learning rate)
\REQUIRE $\beta_1,\beta_2$: exponential decay rates 
\REQUIRE $\theta_0$: initial parameters

\STATE  $m_0\leftarrow0, v_0\leftarrow0, t\leftarrow0$
\STATE  ${\color{NavyBlue}c\leftarrow 2 \cdot (1- |\mu - 0.5|)}$

\WHILE{not converged}

\STATE $t\leftarrow t+1$
\STATE $g_t\leftarrow \nabla_\theta f_t(\theta_{t-1})$
\STATE $m_t\leftarrow \beta_1m_{t-1}+(1-\beta_1)g_t$
\STATE $v_t\leftarrow \beta_2v_{t-1}+(1-\beta_2)({\color{NavyBlue}c \cdot ((1-\mu)g_t - \mu m_t)})^2 +\epsilon$ 
\STATE $\hat{m}_t= m_t/(1-\beta_1^t)$, $\hat{v}_t=v_t/(1-\beta_2^t)$
\STATE $\theta_t\leftarrow \theta_{t-1}-\alpha(\hat{m}_t/\sqrt{\hat{v}_t})$

\ENDWHILE

\end{algorithmic}
\end{algorithm}

We will now calculate $S$ for different values of $\mu$, which allows us to see the relation between AdaFamily and Adam/AdamW, AdaBelief and AdaMomentum. We furthermore will denote with $AdaFamily_{(\mu)}$ the respective variant of the AdaFamily algorithm for a specific $\mu$.
\begin{itemize}
  \item For $\mu=0.0$, $S$ is equal to {\color{NavyBlue}$g_t$}. So the variant $AdaFamily_{(0.0)}$ is similar (but not identical) to the \emph{Adam/AdamW} algorithm.
  \item For $\mu=0.5$, $S$ is equal to {\color{NavyBlue}$g_t-m_t$}. So the variant $AdaFamily_{(0.5)}$ is similar (but not identical) to the \emph{AdaBelief} algorithm.
  \item For $\mu=1.0$, $S$ is equal to {\color{NavyBlue}$-m_t$}. So the variant $AdaFamily_{(1.0)}$ is similar (but not identical) to the \emph{AdaMomentum} algorithm.
  \item For all other values of $\mu$, the variant $AdaFamily_{(\mu)}$ can be seen as a "blend" (mixture) of different Adam-variants. For example, $AdaFamily_{(0.25)}$ can be interpreted as a blend of Adam and AdaBelief, with both contributing "equally" in some way to the blend. In the same way,  $AdaFamily_{(0.75)}$ can be seen as a blend of AdaBelief and AdaMomentum. 
\end{itemize}

The reason why a specific AdaFamily variant is similar, but not exactly identical, to Adam and its variants is because the constant $\epsilon$ is added in different places in the respective algorithms. The placement of $\epsilon$ has an influence \cite{Wang2021} on the algorithm performance.  For AdaFamily, we follow the procedure proposed (and justified) in AdaMomentum and add $\epsilon$ in the preconditioning term $v_k$. 

One can see that with AdaFamily we get an infinite amount of variants of Adam-like algorithms, parametrized via the hyperparameter $\mu$. The hyperparameter $\mu$ determines how "close" the respective AdaFamily variant is to either Adam/AdamW, AdaBelief or AdaMomentum. In section \ref{sec:experiments}, we will evaluate a couple of AdaFamily variants and give guidelines how to set $\mu$ in a proper way.

\section{Experiments and evaluation}
\label{sec:experiments}

We evaluate our proposed AdaFamily algorithm on the task of image classifcation. We compare the AdaFamily variants for $\mu \in \{0.0, 0.25, 0.5, 0.75, 1.0\}$ against the Adam \cite{Kingma2015AdamAM}, AdamW \cite{Loshchilov2019}, AdaBelief \cite{Zhuang2020} and AdaMomentum \cite{Wang2021} algorithm. We employ three standard datasets for image classification: SVHN \cite{Netzer2011}, \mbox{CIFAR\nobreakdash-10} and CIFAR\nobreakdash-100 \cite{Krizhevsky2009}. The datasets consist of 32x32 pixel RGB images, which belong to either 10 classes (SVHN, CIFAR\nobreakdash-10) or 100 classes (CIFAR\nobreakdash-100). For all algorithms, learning rate $\alpha$ is set to the default value $10^{-3}$ and weight decay is set to $10^{-4}$. For all algorithms except for Adam, decoupled weight decay is employed (as proposed in AdamW). The exponential decay parameters are set to their defaults $\beta_1=0.9$ and $\beta_2=0.999$ for all algorithms. The constant $\epsilon$ is set to its default value $10^{-8}$. The mini-batch size is 128 and training is done for 150 epochs, with the learning rate decayed by a factor of 0.5 at epochs 50 and 100. 
We perform the experiments with four popular neural network models for computer vision, taking into account both standard networks (with medium complexity) as well as light-weight networks designed for mobile devices.
We employ ResNet\nobreakdash-50 \cite{He2016DeepRL} and DenseNet\nobreakdash-121 \cite{Huang2017DenselyCC} as standard network models, whereas MobileNetV2 \cite{Sandler2018} and EfficientNet\nobreakdash-B0 \cite{Tan2019} are taken as representatives of light-weight networks. To measure the performance of a certain algorithm, we utilize the top-1 classification error of the final trained model on the test set (which of course has not been seen during training). We do 10 different runs with random seeds and take the average of these 10 runs.

\newcommand{\JrsAlignNr}[1]{\multicolumn{1}{S[detect-weight,table-format=2.2]}{#1}}
\newcommand\JrsBlue{\color{NavyBlue}} 
\newcommand\JrsRed{\color{Maroon}}

\begin{table}[b]
\small
\centering
\caption{Results (measured as test error, in percent) on the \emph{CIFAR-10} dataset for different models. The best and second-best result for each model is marked in {\color{Maroon}red} and {\color{NavyBlue}blue}. }
\label{tbl:cifar10}
\begin{tabular*}{0.85\textwidth}{@{\extracolsep{\fill}} l *{4}{S[detect-weight,table-format=2.2]} }
\toprule
\textbf{Algorithm} & {ResNet-50} & {DenseNet-121} & {MobileNetV2} & {EfficientNet-B0} \\
\midrule
Adam                 &           12.89 &          10.31 &           14.33 & 21.18 \\
AdamW                &           13.27 &           9.32 &           15.18 & 21.41 \\
AdaBelief            &           12.70 &           8.93 &           14.97 & 21.45 \\
AdaMomentum          &           14.11 &           9.48 & \JrsRed   14.15 & \JrsBlue 19.56 \\
AdaFamily$_{(0.0)}$  & \JrsBlue  12.69 &           8.93 &           15.07 & 21.61 \\
AdaFamily$_{(0.25)}$ &           12.71 & \JrsRed   8.89 &           15.34 & 22.29 \\
AdaFamily$_{(0.5)}$  & \JrsRed   12.65 & \JrsBlue  8.92 &           14.85 & 21.55 \\
AdaFamily$_{(0.75)}$ &           13.79 &           9.21 &           14.19 & \JrsRed 19.36   \\
AdaFamily$_{(1.0)}$  &           14.56 &           9.50 & \JrsBlue  14.18 & 19.67 \\
\bottomrule
\end{tabular*}
\end{table}

The results of the experiments are shown in Table \ref{tbl:cifar10} for CIFAR-10, Table \ref{tbl:cifar100} for CIFAR-100 and Table \ref{tbl:svhn} for SVHN dataset. 
One can observe that the variants of the AdaFamily algorithm are outperforming Adam/AdamW, AdaBelief and AdaMomentum in most cases. For all datasets and all network models, a specific variant of AdaFamily is either the best-performing or second-best algorithm.

Regarding how to choose the hyperparameter $\mu$ for the AdaFamily algorithm, one can draw the conclusion that smaller values of $\mu$ perform better for standard models (ResNet-50, DenseNet-121), whereas larger values of $\mu$ perform better for light-weight models (MobileNetV2, EfficientNet-B0). Based on this, a rough guideline could be to employ the variant $AdaFamily_{(0.25)}$ for standard models (with medium complexity) and $AdaFamily_{(0.75)}$ for light-weight models. As mentioned in the description of AdaFamily algorithm, $AdaFamily_{(0.25)}$ can be interpreted as a blend of Adam/AdaBelief, whereas $AdaFamily_{(0.75)}$ can be seen as a blend of AdaBelief/AdaMomentum.

\begin{table}[t]
\small
\centering
\caption{Results (measured as test error, in percent) on the \emph{CIFAR-100} dataset for different models. The best and second-best result for each model is marked in {\color{Maroon}red} and {\color{NavyBlue}blue}. }
\label{tbl:cifar100}
\begin{tabular*}{0.85\textwidth}{@{\extracolsep{\fill}} l *{4}{S[detect-weight,table-format=2.2]} }
\toprule
\textbf{Algorithm} & {ResNet-50} & {DenseNet-121} & {MobileNetV2} & {EfficientNet-B0} \\
\midrule
$Adam$ &               41.11 & 32.91 & 42.84 & 52.55 \\
$AdamW$ &              39.78 & 31.55 & 44.88 & 53.13 \\
$AdaBelief$ &          40.15 & 29.40 & 42.66 & 51.85 \\
$AdaMomentum$ &        43.72 & 30.37 & \color{NavyBlue}40.20 & 50.58 \\
$AdaFamily_{(0.0)}$  & \color{NavyBlue}39.05 & \color{NavyBlue}29.39 & 43.09 & 52.75 \\
$AdaFamily_{(0.25)}$ & \color{Maroon}38.81 & \color{Maroon}29.21 & 43.37 & 53.25 \\
$AdaFamily_{(0.5)}$  & 39.43 & 29.46 & 42.71 & 52.05 \\
$AdaFamily_{(0.75)}$ & 42.05 & 30.15 & 40.47 & \color{Maroon}50.17 \\
$AdaFamily_{(1.0)}$  & 43.35 & 30.32 & \color{Maroon}40.06 & \color{NavyBlue}50.31 \\
\bottomrule
\end{tabular*}
\end{table}

\begin{table}[t]
\small
\centering
\caption{Results (measured as test error, in percent) on the \emph{SVHN} dataset for different models. The best and second-best result for each model is marked in {\color{Maroon}red} and {\color{NavyBlue}blue}. }
\label{tbl:svhn}
\label{tbl:cifar100}
\begin{tabular*}{0.85\textwidth}{@{\extracolsep{\fill}} l *{4}{S[detect-weight,table-format=2.2]} }
\toprule
\textbf{Algorithm} & {ResNet-50} & {DenseNet-121} & {MobileNetV2} & {EfficientNet-B0} \\
\midrule
$Adam$ &               4.86 & 4.55 & 4.88 & 6.64 \\
$AdamW$ &              4.30 & 3.82 & 4.58 & 6.25 \\
$AdaBelief$ &          4.18 & \color{NavyBlue}3.59 & 4.53 & 6.23 \\
$AdaMomentum$ &        4.71 & 3.79 & 4.52 & \color{NavyBlue}6.10 \\
$AdaFamily_{(0.0)}$  & \color{NavyBlue}4.17 & 3.63 & 4.56 & 6.47 \\
$AdaFamily_{(0.25)}$ & \color{Maroon}4.17 & \color{Maroon}3.58 & 4.62 & 6.40 \\
$AdaFamily_{(0.5)}$  & 4.18 & 3.67 & 4.57 & 6.30 \\
$AdaFamily_{(0.75)}$ & 4.45 & 3.73 & \color{Maroon}4.51 & \color{Maroon}6.01 \\
$AdaFamily_{(1.0)}$  & 4.58 & 3.76 & \color{NavyBlue}4.51 & 6.21 \\
\bottomrule
\end{tabular*}
\end{table}


Note that AdaFamily does not increase the memory consumption of the training (compared to Adam and its variations), as no additional vectors have to be calculated and stored. This is especially important when porting the algorithm to a \emph{distributed training} framework, where usually large models (e.g. transformer models for NLP) are trained. It has also practically the same runtime, as no significant additional computations are performed.

\section{Conclusion and future work}
\label{sec:conclusion}

We presented \emph{AdaFamily}, a family of adaptive gradient methods parametrized by hyperparameter $\mu$ which can be interpreted as sort of a \emph{blend} of the optimization algorithms Adam, AdaBelief and AdaMomentum. Experiments on the task of image classification demonstrated that our proposed method outperforms these algorithms in most cases.

In the future, we will port the AdaFamily algorithm to a distributed training framework like \emph{DeepSpeed}. The combined formulation allows for a faster porting of the algorithm to a distributed training framework, as only one Algorithm (AdaFamily) has to be ported instead of multiple ones.

Furthermore, we would like to experiment with varying the hyperparameter $\mu$ during training (instead keeping it constant as it is done now). For example, one could vary $\mu$ linearly from $\mu=0.0$ at the begin of training to $\mu=0.5$ at the end of the training. By this, we are kind of \emph{morphing the employed optimizer during the training process}.

Finally, it could be useful to investigate the properties of AdaFamily more in depth, by analyzing its convergence properties and the like. This could help also to explain why a smaller $\mu$ seems to be better for standard models and a higher $\mu$ for light-weight models.


\section*{Acknowledgment}
This work was supported by European Union´s Horizon 2020 research and innovation programme under grant number 951911 - AI4Media.

%
%
%
\bibliographystyle{splncs04}
\bibliography{bibliography}

\end{document}